\documentclass{article}

\PassOptionsToPackage{numbers, compress}{natbib}
\usepackage[preprint]{neurips_2024}

\usepackage[utf8]{inputenc} 
\usepackage[T1]{fontenc}    
\usepackage[colorlinks,
            linkcolor=red,
            anchorcolor=blue,
            citecolor=green
            ]{hyperref}
\usepackage{url}            
\usepackage{booktabs}       
\usepackage{amsfonts}       
\usepackage{nicefrac}       
\usepackage{microtype}      
\usepackage{xcolor}         
\usepackage{caption}
\usepackage{subcaption}
\usepackage{natbib}
\usepackage{graphicx}
\usepackage{enumitem}
\usepackage{makecell}
\usepackage{amsmath}
\usepackage{amsthm}
\usepackage{thmtools}
\usepackage{thm-restate}
\usepackage{algorithm}
\usepackage{algorithmic}
\usepackage{bbm}
\usepackage{tabularx}
\usepackage{listings}
\usepackage{xcolor}
\usepackage{multirow}
\usepackage{multicol}
\usepackage{tcolorbox}
\usepackage{tikz}
\usepackage{footnote}
\usepackage{wrapfig}

\definecolor{tblue}{RGB}{31,119,180}
\definecolor{torange}{RGB}{255,127,14}
\definecolor{tgreen}{RGB}{44,160,44}
\definecolor{tred}{RGB}{214,39,40}
\definecolor{tpurple}{RGB}{148,103,189}
\definecolor{lightblue}{RGB}{173, 216, 230}
\definecolor{lightpink}{RGB}{255, 182, 193}
\definecolor{lightgreen}{RGB}{144, 238, 144}

\usepackage{colortbl}
\usepackage{xcolor}
\usepackage{array}

\newcommand{\hide}[1]{} 

\def\model{TacticExpert}

\title{TacticExpert: Spatial-Temporal Graph Language Model for Basketball Tactics}

\author{
  Lingrui Xu\textsuperscript{1} ~~~
  Mandi Liu\textsuperscript{2} ~~~
  Lei Zhang\textsuperscript{1}\thanks{Lei Zhang is the Corresponding Author.} \\
  \textsuperscript{1}Beijing Jiaotong University ~~~
  \textsuperscript{2}Carnegie Mellon University \\
  \texttt{\{20711029, zhlei\}@bjtu.edu.cn} \\
  \texttt{ballonmandy@outlook.com}
}

\begin{document}

\maketitle

\begin{abstract}

The core challenge in basketball tactic modeling lies in efficiently extracting complex spatial-temporal dependencies from historical data and accurately predicting various in-game events. Existing state-of-the-art (SOTA) models, primarily based on graph neural networks (GNNs), encounter difficulties in capturing long-term, long-distance, and fine-grained interactions among heterogeneous player nodes, as well as in recognizing interaction patterns. Additionally, they exhibit limited generalization to untrained downstream tasks and zero-shot scenarios. In this work, we propose a Spatial-Temporal Propagation Symmetry-Aware Graph Transformer for fine-grained game modeling. This architecture explicitly captures delay effects in the spatial space to enhance player node representations across discrete-time slices, employing symmetry-invariant priors to guide the attention mechanism. We also introduce an efficient contrastive learning strategy to train a Mixture of Tactics Experts module, facilitating differentiated modeling of offensive tactics. By integrating dense training with sparse inference, we achieve a 2.4x improvement in model efficiency. Moreover, the incorporation of Lightweight Graph Grounding for Large Language Models enables robust performance in open-ended downstream tasks and zero-shot scenarios, including novel teams or players. The proposed model, \model, delineates a vertically integrated large model framework for basketball, unifying pretraining across multiple datasets and downstream prediction tasks. Fine-grained modeling modules significantly enhance spatial-temporal representations, and visualization analyzes confirm the strong interpretability of the model. Uniquely, the \model introduces geometric deep learning into the messaging of the spatio-dependent coding layer of the Graph Transformer architecture to model four second-order dihedral group symmetry views of the basketball court, namely identity transformation, rotation 180 degrees around the X-axis, rotation 180 degrees around the Y-axis, and rotation 180 degrees around the z-axis. Specifically, we propose equivariant attention mechanism that inputs four symmetric views enhanced by the initial spatiotemporal data embedding layer into the spatiotemporal dependent coding layer at one time and allows these views to explicitly interact in a manner that does not break symmetry. Through symmetric constraints on the message passing layer, the output of the network is consistent with the symmetric transformation of the input.

\end{abstract}

\clearpage

\section{Introduction}\label{sec:intro}

Basketball is a physically competitive sport that is focused on shooting and is highly dependent on strategic teamwork, which makes it very tactical. Modeling basketball tactics from a spatial-temporal perspective by analyzing player trajectories with event types and locations across sequential time slices, providing a natural representation of in-game scenarios~\cite{liu2022motion}. Early research was mainly based on shallow probabilistic statistical models. Miller et al. proposed a heuristic method that combines spatial priors and non-negative matrix factorization to construct a generative latent factor model for analyzing and predicting shot selection~\cite{miller2014factorized} used Gaussian processes for spatial regularization to estimate the probability of the next game event under the current game state~\cite{yue2014learning} attempted to identify spatial patterns that improve team performance in offense and defense by modeling changes in the convex hull area formed by the five players on the court using a dual-state hidden Markov model, linking these changes to team performance, on-court lineups, and tactical strategies~\cite{metulini2018modelling} evaluated the value of different court regions by simulating the ball movement during possession changes to identify teams capable of effectively suppressing opponents’ control of high-value regions~\cite{cervone2016nba} used a Bayesian hierarchical model to estimate the shooting accuracy of the players in each position on the half court, compared the estimated shooting accuracy with actual shooting attempts and provided the optimal shooting distribution in a given lineup, comparing the actual total scores with the suggested optimal total scores~\cite{sandholtz2020measuring} constructed a rebound model based on players’ shooting attempt locations, the ball’s rebounding location, and the height of the ball during the rebound. The features extracted from the model were used to train a binary classifier to predict whether the offensive team would successfully secure the rebound~\cite{masheswaran2014three}.

Further research has improved the ability of models to perceive tactical patterns by explicitly modeling basketball tactics. Franks used a hidden Markov model to identify matchups in man-to-man defensive tactics, representing the average positions of the defensive players in a given time slice as a convex combination of the positions of the offensive players, the ball position and the basket position. The defensive players' positions were modeled as a Gaussian distribution around the average positions given the observed positions of the offensive players~\cite{franks1986eyewitness} designed a discriminative temporal interaction network to characterize group movement patterns in basketball, learning multi-modal densities for predefined group movement patterns using a well-defined Riemannian metric and identifying the most likely game scenario given an input set of trajectories using an MAP classifier~\cite{li2009learning}. identified tactics by analyzing frequently occurring passing sequences in games, using detailed node information on the passer and receiver and the pass reception location as features to recognize tactics consistent in participating players and key locations~\cite{wang2015discerning}. Recently, advances in tracking technology have significantly improved the accessibility and quality of spatial-temporal data in basketball, enabling more precise descriptions of on-court events. Technologies such as Global Positioning Systems (GPS)~\cite{catapult} and Radio Frequency Identification (RFID)~\cite{skinner2010price} employ optical devices attached to players’ uniforms or embedded within the ball to capture motion trajectories. The SportVU system of the National Basketball Association (NBA) uses six motion capture cameras located around the court, recording at 25 frames per second to generate time-stamped coordinates (x, y) for the players and more than 20 frames per second for the basketball coordinates (x, y, z). The latest high-resolution systems achieve coordinate accuracy within one centimeter~\cite{terner2021modeling}.

As more and more industry vendors have been providing deployment services for advanced tracking devices to professional teams and leagues, it is possible to apply rich, high-quality spatio-temporal data for basketball tactical analysis. Unlike typical topics in the field of spatiotemporal data mining, such as crime prediction and traffic flow prediction, spatiotemporal motion trajectories extracted from basketball games usually exhibit characteristics such as smaller temporal and spatial scales, dense sampling rates, a limited number of player nodes, and interactions between players characterized by cooperation and opposition~\cite{gudmundsson2017spatio}. These characteristics result in problems such as too few nodes, sparse interactions between nodes, and heterogeneity in relationships between adjacent nodes in the abstracted graphs where players are represented as nodes. Due to these issues, most of the early work on basketball tactics modeling focused on relatively small datasets~\cite{goldsberry2013dwight}, did not construct predictive models, used coarse aggregate statistics that did not simulate specific game scenarios or tactics~\cite{miller2014factorized}, and did not consider the context of opposing teams~\cite{bialkowski2014win}. Moreover, a significant amount of manual annotation was required~\cite{lowe2013lights}.

Since dividing datasets by basketball tactics results in large data volumes, recent work has increasingly focused on building deep models based on GNNs (graph neural networks) and their variants to model larger-scale data scenarios. TacticAI~\cite{wang2024tacticai} designed equivariant graph convolutional neural networks to aggregate the features of the two-hop neighbor player node, enabling representation updates for target players. Nistala et al. used a graph convolutional auto-decoder to extract continuous features of a series of movement routes and then applied k-means clustering for unsupervised tactic recognition~\cite{nistala2019using}.

Overall, current research has shown high predictive accuracy in downstream tasks by modeling the motion trajectories and event logs of real game scenarios, but it remains limited in long-term, long-range, and spatio-temporal dynamic interaction tasks. The end-to-end training paradigm restricts the generalization capabilities of the models. However, with the increasing availability of spatio-temporal data in basketball, finer-grained modeling of basketball tactics is becoming increasingly feasible. The primary challenge in spatial-temporal basketball tactic modeling lies in effectively extracting dynamic and intertwined spatial-temporal dependencies from historical tactical data. This paper explores existing approaches, their limitations, and introduces a novel method to address these challenges in basketball tactic modeling.

The remainder of the article is organized as follows. In Section~\ref{sec:review}, we discuss the emerging development of techniques that combine graph learning with large language models (LLMs) over the past years. Section~\ref{sec:preliminary} introduces the preliminaries and formalizes the problem studied. Section~\ref{sec:methodology} proposes the method for making meaningful and context-aware predictions for basketball games. In Section~\ref{sec:experiments},  we conduct extensive experiments to evaluate the performance of our method. The concluding remarks are given in Section~\ref{sec:conclusion}.

\section{Literature Review}\label{sec:review}

\subsection{Large language model as representation enhancers}

Inspired by the emerging capabilities, superior understanding and reasoning abilities exhibited by LLMs after large-scale text pre-training, as well as successful practices in computer vision—such as DALL·E2~\cite{openai2024} and Llava~\cite{liu2024visual}—which build multimodal models by fine-tuning LLMs, this field has seen significant growth and innovation.

Early attempts at using large language models (LLMs) for graph learning began with enhancing textual representations in rich-text graphs, such as academic citation networks and e-commerce product co-purchase networks. Early work like TAPE~\cite{he2023harnessing} involved inputting node representations and connection relationships from sampled subgraphs of academic citation networks into a large model in text form. The model outputted node classification results and provided reasoning for its judgments, thereby enhancing the initial representation data. GNNs were then used to generate node embeddings through ensemble learning for downstream tasks. This work demonstrated that merely using LLMs to enhance textual data could surpass traditional, complex graph representation learning models on the OGB-arXiv leaderboard and achieve new state-of-the-art (SOTA) performance, showcasing the potential of combining LLMs with graph learning.

This approach was later extended to the recommendation system domain, where it was used to augment user and product profile features. LLMRec~\cite{wei2024llmrec} input user and product node information into a large model to re-describe profile information and generate positive and negative sample pools, addressing issues like product exposure sparsity and noisy data in recommendations. RLMRec~\cite{ren2024representation} used LLMs to integrate auxiliary textual signals to capture complex semantic information about user behaviors and preferences. It also adopted a cross-view alignment framework to align the semantic space of the LLM with the representation space of collaborative relationships, significantly outperforming traditional benchmark models in the recommendation field, such as LightGCN~\cite{he2020lightgcn} and HGCF~\cite{sun2021hgcf}.

Additionally, LLMs as enhancers can improve the understanding of label categories in graph datasets, alleviating data sparsity issues in node classification tasks with missing labels by generating pseudo-labels. They also map different label sets into the same semantic space, thereby enhancing the model generalization capabilities. LLM-GNN~\cite{chen2023label} used LLMs to generate pseudo-labels for unlabeled nodes based on their textual features. SimTeG~\cite{duan2023simteg} adopted a two-stage training process, first fine-tuning an LLM to generate node embeddings and then training a GNN on these embeddings, simplifying text classification and text similarity tasks.

Using LLMs as enhancers involves transferring their understanding and generalization capabilities from text to the textual features of traditional representation learning models. By integrating and complementing these enhanced features with representations encoded by graph encoders, more accurate predictions for downstream tasks can be achieved. LLMs are used not only as text encoders for feature extraction but also to leverage their exceptional text-to-text capabilities to further refine and enrich raw textual features. This highlights key points, provides explanations and reasoning, and enriches domain-specific knowledge.

Moreover, introducing the unified textual representation space of LLMs allows for generalization across datasets, even when the textual feature sets differ completely, as LLMs represent features within a unified space. However, using LLMs as representation enhancers has certain limitations. Experiments have found, counterintuitively, that stronger models are not always better for data augmentation. Decoder-only models like GPT series, Llama, and Claude do not perform as well as simpler pre-trained models with retrieval capabilities, such as sentence-BERT~\cite{chen2024exploring}. Furthermore, while LLMs improve representation quality, they do not necessarily enhance the performance of GNNs, indicating a performance ceiling.

\subsection{Large language model as predictors}

Using large models as predictors involves leveraging them as the underlying architecture for making predictions on graph-related tasks. Typically, this approach converts graph-structured data into text format, directly describing the structure and features of the graph through text or transforming the graph structure into latent representations. These latent representations are aligned with the textual descriptions processed by the language model. Given the exceptional generalization capabilities of large language models (LLMs) on textual data, this method has potential for transferability across different downstream tasks involving graph data. The key lies in how the graph structure and feature information are represented in the input prompts to the LLM, enabling the model to comprehend the graph.

The NLGraph benchmark~\cite{wang2024can} explores LLM's ability to understand graphs through a question-answer format. It inputs nodes, node meanings and connection relationships of a graph as text into the LLM, testing its ability to understand and reason about structured data, for example, determining the existence of loops, finding the shortest paths between nodes, or modeling changes in representations after simulating one layer of GNN message propagation and aggregation. The benchmark also incorporates techniques like Chain of Thought (CoT) and algorithmic prompts. The results indicate that LLMs have a basic ability to perform graph reasoning but exhibit significant shortcomings, such as adverse effects of advanced prompt techniques on complex problems. Furthermore, LLM predictions can show unstable fluctuations caused by seemingly trivial statistical information. This approach is constrained not only by the token length limitation of LLM inputs but also by the inability of LLMs to pass graph-specific tests, such as two-hop coloring, demonstrating a lack of understanding of graph structure information described in text.

Subsequent models addressed these issues by introducing custom tokens in prompts to replace textual descriptions of graph structures. Early work like GPT4Graph~\cite{guo2023gpt4graph} inputs target nodes and their one-hop neighbors into the LLM, emulating the GNN process of aggregating neighbor information. The model summarized the content of the input subgraph and generated a \texttt{<text>} token to represent the aggregated information, which was then input along with node representations for reasoning. The experiments showed strong generalization capabilities in zero-shot scenarios, but the imitation of GNN operations for neighbor information aggregation minimally contributed to predictive performance. Furthermore, this approach requires extensive textual information for nodes, which is often lacking in graphs such as protein molecules or session-based recommendation networks.

GraphText~\cite{zhao2023graphtext}, building on this framework, mitigated the need for textual information from nodes by generating cluster labels for nodes using LLM. However, this led to significant information loss and was only effective in few-shot scenarios. More recent approaches abandoned the effort to make LLMs directly understand graph structure information. Instead, they used GNN-like models, which are better suited to encode graph structure, to generate embeddings and integrated the results into prompts as \texttt{<graph>} tokens. Aligning these \texttt{<graph>} tokens with the textual representation space of the LLMs enabled the understanding of the information of the graph structure.

For example, GraphGPT~\cite{tang2024graphgpt} and LLaGA~\cite{chen2024llaga} trained simple linear mappers to align the graph modality with the text modality space. Specifically, CLIP-based contrastive alignment between textual encoding and graph structure encoding allowed graph encoders to generate node embeddings in the same semantic space as LLM textual representations. These embeddings were then input into the LLM as \texttt{<graph>} tokens, with self-supervised and downstream task-specific fine-tuning enabling the LLM to understand the \texttt{<graph>} tokens. The experiments demonstrated improved inductive biases and understanding of graph structures while significantly reducing the token length of input prompts. GraphTranslator~\cite{zhang2024graphtranslator} further refined alignment modules with BLIP-2 dual-tower models for more granular integration.

Using LLMs as predictors represents a comprehensive and feasible approach to integrating large models into graph learning.

\subsection{Graph foundation models based on large models}

Inspired by foundational models in language, vision, and multimodal domains—where unified architectures are pre-trained on extensive datasets with self-supervised tasks and fine-tuned for diverse downstream tasks—we aim to build a unified foundational architecture for graphs. This architecture seeks to address all tasks on any graph, leveraging inductive biases from different graphs to achieve strong performance in zero-shot or few-shot scenarios, such as long-tail predictions, out-of-distribution generalization, and cold-start problems.

Analogously to the tokenization process in large language models, a potential approach involves identifying the smallest "graph vocabulary units" in graph-structured data. Based on a divide-and-conquer strategy, large graphs can be segmented into graph vocabulary units, computed, and aggregated to handle arbitrary graph inputs. Current research on constructing foundational graph models is in its early stages, with researchers exploring expressive and structure-preserving graph vocabulary units. Word embeddings optimized in large language models serve as an intermediate medium for storing information about various graph vocabulary units.

The One for All model~\cite{liu2023one} leverages the understanding capabilities of large models for textual graph features to achieve generalization across graphs. It also proposes a unified graph prediction task format, enabling different graph prediction tasks, such as node classification, graph classification, and link prediction, to be performed using a single generalized task format. This approach allows training on one task to generalize across multiple graph tasks.

A more comprehensive model, Ultra~\cite{metulini2018modelling}, eliminates the dependency on large language models' text generalization capabilities and pre-trained textual representations. It proposes a foundational model based on knowledge graphs, cleverly transforming universal head-to-head, tail-to-tail, and head-to-tail transfer relationships into semantic representations. This enables a common representation method across different knowledge graphs.

OpenGraph~\cite{xia2024opengraph} improves generalization across different input graphs by smoothing graph signals through higher-order multiplication of adjacency matrices. Subsequently, a large language model is used to identify corresponding semantic information for the smoothed graph units.

These early efforts highlight the promise of foundational models in the graph domain, aiming to replicate the success of foundational architectures in other modalities.

\subsection{Spatial-temporal Data Mining Techniques Based on Deep Learning}

Spatiotemporal data mining aims to identify spatiotemporal dependency patterns by analyzing historical spatiotemporal data and learning the mapping function from past spatiotemporal events to future ones, thereby enabling downstream prediction tasks. The fundamental challenge lies in effectively capturing and modeling the dynamic and intertwined spatio-temporal relationships within the data.

Early research modeled spatial and temporal dependencies using separate approaches. Zhang et al. first divided the target spatial domain into grids and used convolutional neural networks (CNNs) based on domain aggregation to model the spatial dimension, while employing recurrent neural networks (RNNs), which excel in sequential modeling, for the temporal dimension~\cite{zhang2017deep}. To address the catastrophic forgetting problem of RNN, Yao et al. later adopted long-short-term memory (LSTM) neural networks to optimize temporal modeling~\cite{yao2018deep}.

Since raw spatial points of interest (POI) are naturally abstracted as graph-structured data, models based on graph neural networks (GNNs) have been extensively applied to spatial dependency modeling, leading to the emergence of spatiotemporal graph neural networks (STGNNs). The STGNN framework is flexible, with various studies proposing different models to design message-passing mechanisms between nodes, which can be categorized into spatial attention modeling and temporal attention modeling.

For spatial attention,~\cite{fernando2018soft+} proposed a composite attention model that uses "soft attention" and "hard-wired" attention to map the trajectory information from local neighbors to future positions of target pedestrians. For temporal attention, DeepCrime~\cite{huang2018deepcrime} introduced a hierarchical attention recurrent network model for crime prediction. The temporal attention mechanism captures the relevance of crime patterns learned from previous time periods to assist in predicting future crimes, and automatically assigns importance weights to hidden states learned from different time slices. Similarly,~\cite{liang2018geoman} proposed a multilevel attention network to predict geospatial time series generated by sensors deployed at different geographic locations, allowing continuous and collaborative environmental monitoring, such as air quality. PDFormer~\cite{jiang2023pdformer} used a spatio-temporal self-attention mechanism to model periodicity and local spatial dependencies in traffic flow data.

However, most existing research still focuses on capturing dependencies within a single dimension using attention mechanisms, leaving room for improvement in modeling the intricate interactions between spatial and temporal dimensions.

\section{Preliminary}\label{sec:preliminary}

\begin{figure*}[t]
    \centering
    \includegraphics[width=\columnwidth]{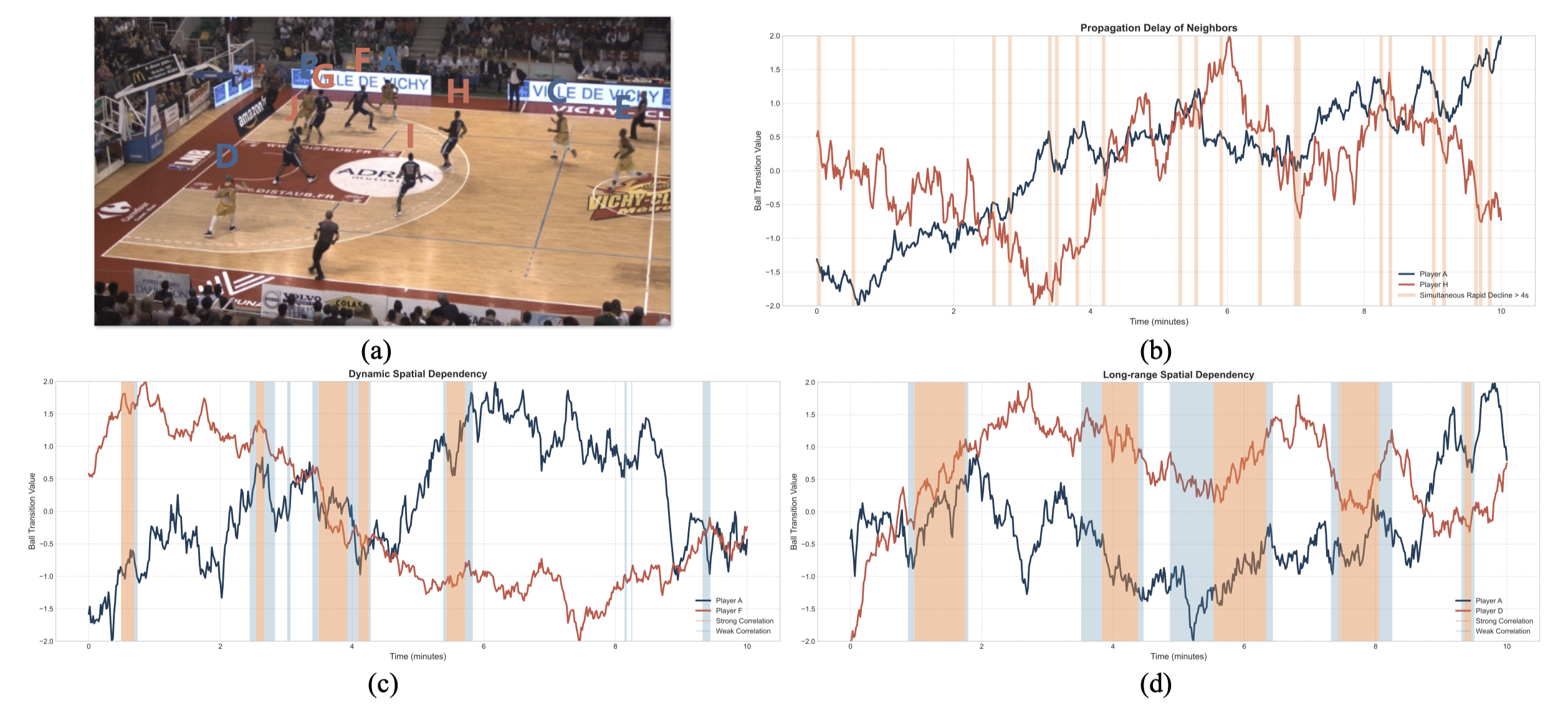}
    \caption{The findings in basketball scenarios. (a) Player nodes distribution (blue and orange represent the two teams); (b) Propagation delay effect of neighbor nodes (using A and H as examples); (c) Spatial dynamic dependency of player nodes (using A and F as examples); (d) Long-range spatial dependency of player nodes (using A and D as examples).}
    \label{fig:finding}
\end{figure*}

\subsection{Definition ball transition tensor}
We denote the transition of the ball at time t for all nodes of the player $N$ on the player position graph by \(\mathbf{X}_t \in \mathbb{R}^{N \times C}\), where $C = -1$ indicates a pass, $C = 1$ indicates receiving the ball, and $C = 0$ indicates neither receiving nor passing the ball, or continuously holding the ball. We use the three-dimensional tensor \(\mathbf{X} = (X_1, X_2, \dots, X_T) \in \mathbb{R}^{T \times N \times C}\) to denote the ball transition tensor of all player nodes over a total of $T$ time slices.

\subsection{Problem statement}

\textbf{Encoder}. Given the historically observed ball transition tensor X, the goal of the encoder block in \model is to uncover the spatial-temporal dependency patterns through fine-grained modeling of the basketball scenarios. This enables learning a mapping function \(f(\theta_{\text{Encoder}})\) from the previous observation values to predict the ball transition of the player nodes for the next time step through representation learning. The initial embedding of the feature of the player nodes at time t is denoted as \(F_t\).

\[
\theta^*_{\text{Encoder}} = \arg \max_{\theta_{\text{Encoder}}} \mathbb{E}_{p(X, \mathbf{F})} \left[ p(X_{t+1} \mid X_t, \mathbf{F}_t; \theta_{\text{Encoder}}) \right]
\]

\begin{figure*}[t]
    \centering
    \includegraphics[width=\textwidth]{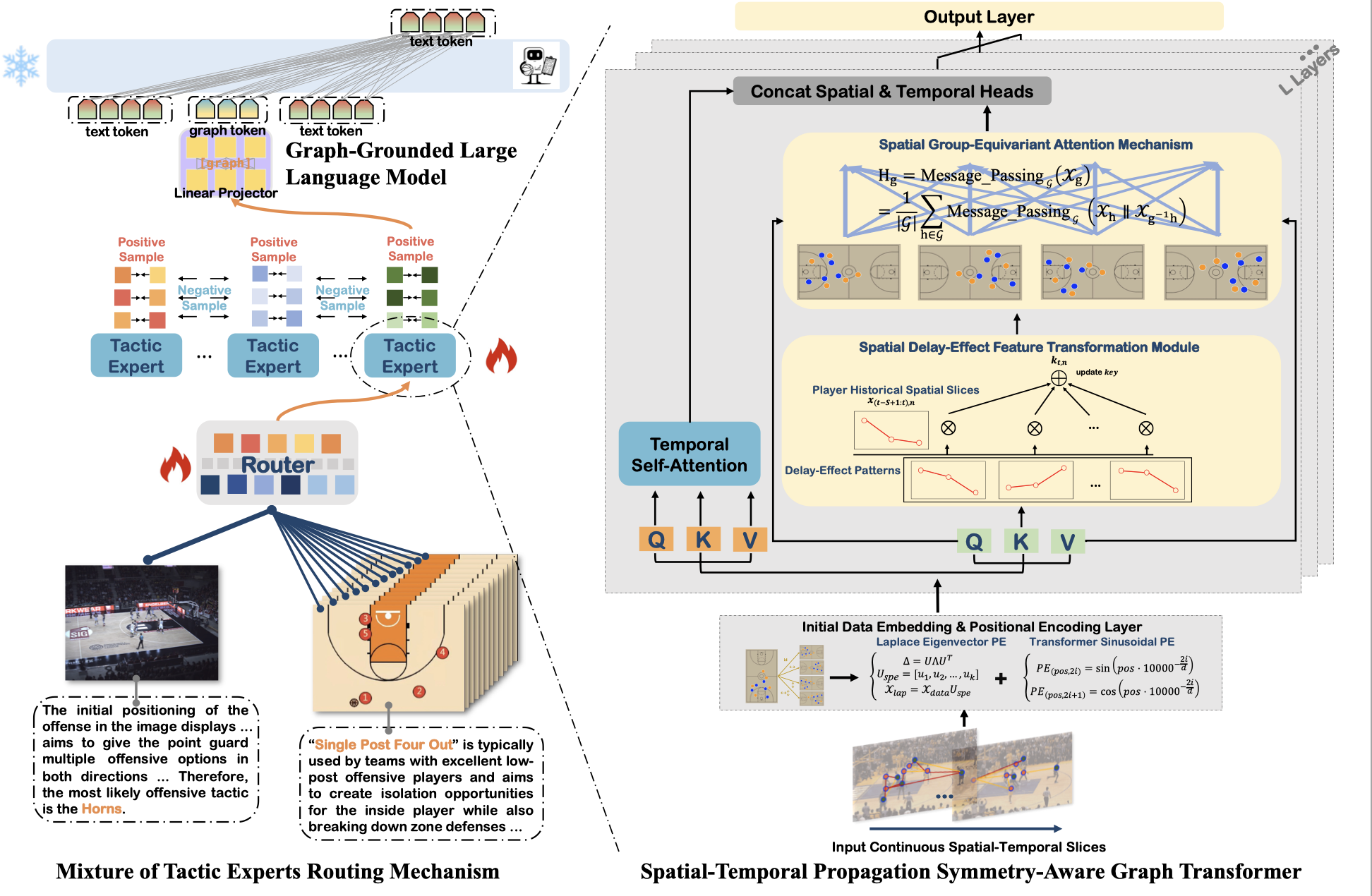}
    \caption{The overall architecture of \model.}
    \label{fig:framework}
\end{figure*}

\textbf{Decoder}. The decoder block is designed to align the player node embeddings generated by the encoder block with the representation space of LLMs. By incorporating these embeddings as graph tokens alongside textual context prompts, the decoder inputs them into the LLM to facilitate open-ended downstream task predictions in zero-shot scenarios. The tasks involved in decoder optimization include: (i) node classification task 'who is most likely to hold the ball at a specific moment?' (ii) link prediction task 'will the current ball handler pass the ball?' to whom is she/he most likely to pass? (iii) graph classification task 'will the ball handler attempt a shot at a specific moment?', which requires computing the representations of all player nodes and making a judgment using a readout function. In this setting, the mapping function \(f(\theta_{\text{Decoder-LLM}})\) represents the capabilities acquired by the LLM during pre-training. It does not require additional learning of the target data.
\[
\theta^*_{\text{Decoder}} = \arg \max_{\theta_{\text{Decoder}}} \mathbb{E}_{p(X, \mathbf{F})} \left[ p(X_{(t+1):(t+T')} \mid X_{(t-T+1):t}, \mathbf{F}_{(t-T+1):t}; \theta_{\text{Decoder}}) \right]
\]
Figure \ref{fig:framework} shows the overall architecture of \model, which comprises three key components: a mixture of tactic experts routing mechanism, a Graph Transformer that is sensitive to spatial and temporal symmetry and a lightweight graph grounding module for large language models. In the next section, we will describe each module in detail.

\section{Methodology}\label{sec:methodology}

\subsection{Mixture of tactic experts routing mechanism}

To explicitly account for offensive tactical patterns and mitigate the risk of overfitting due to imbalanced tactical distributions in the dataset, we employ multiple experts to differentiate spatial-temporal predictions under various tactics. Due to the scarcity of domain-specific training data, adopting the conventional mixture-of-experts (MoE) architecture used in large language models is impractical. The standard MoE approach relies on vast pretraining datasets to gradually allow experts to specialize in specific knowledge areas without requiring fine-grained partitioning of the pretraining dataset. This minimizes the injection of human priors and enables the model to autonomously discover hidden patterns. However, such methods may fail to converge with smaller datasets. To address this limitation, \model preemptively injects human prior knowledge by finely partitioning the pretraining dataset, guiding the experts to specialize in certain tactics early in the training process. This strategy facilitates faster convergence on smaller datasets, but can limit the model's capacity to fully realize its potential on larger datasets. Additionally, our modified MoE architecture supports dense training with sparse inference, activating only a subset of model parameters during inference, thereby significantly improving response efficiency. Specifically, we first utilize a pre-trained multi-modal large model to enhance the textual descriptions of continuous spatial-temporal slices in basketball game videos. The enhanced text embeddings are then compared with standard offensive tactics based on semantic similarity. Finally, we employ an efficient and interpretable contrastive learning strategy to train multilayer perceptron (MLP)-routed tactical experts, with each expert independently performing spatial-temporal encoding.

Multi-modal data augmentation and semantic encoding. To route the pretraining datasets to different experts based on the semantic similarity between standard offensive basketball tactics and those employed in real games within the pretraining data, we first collected images of the initial player positions and textual descriptions for 12 standard offensive tactics. We also extracted images corresponding to the initial timestamp (time zero) from the pretraining dataset, representing the start of each tactic. These images and tactic descriptions were input into a multi-modal large model, which leveraged its knowledge for feature augmentation through appropriate prompting. The output was a fixed-format textual description that included the initial player positions, the dynamic progression of the tactic, and the scoring principles. Similarly, we input the initial player position images from the pretraining dataset into the multi-modal model and employed chain-of-thought (CoT) prompting to trigger complex reasoning. This guided the model in inferring the most likely offensive tactic from the 12 standard candidates and providing an interpretative analysis of its inference. Subsequently, using text as the modality, we encoded the augmented textual descriptions of the 12 standard basketball tactics and the textual descriptions of the real-game tactics from the dataset using Sentence-BERT. We then calculated the cosine similarity between the embeddings of the standard basketball tactics and those of the real-game tactics.

Contrastive learning for expert routing. \model leverages the semantic similarity between real games and standard tactics (encoded by pre-trained Sentence-BERT) and adopts a multilayer perceptron (MLP) with two fully connected layers as a tactical expert router. Instead of aggregating weighted spatial-temporal embeddings from multiple experts, it selects the expert with the highest routing weight for encoding each real-game tactic. Consequently, each expert must have a clear tactical preference to ensure routing sparsity and to maintain diverse and comprehensive tactical coverage. Although we introduce human priors by partitioning the pretraining dataset based on tactics, this does not guarantee that experts will learn knowledge aligned with human intent during deep network training. Without constraints, potential issues include: (1) dominance by a few experts, leading to infrequent activation of others, and (2) overlapping expert responsibilities lead to evenly distributed routing weights, with tactical knowledge dispersed across multiple experts. This dispersion can result in significant information loss when only one expert is selected for routing. To address these challenges, \model employs self-supervised contrastive learning to constrain the functional scope of each expert. Embeddings generated by the same expert are positive samples, while those generated by different experts are negative samples. This approach minimizes intra-expert distance and maximizes inter-expert distance, enhancing mutual information within each expert and reducing redundancy across experts, thereby promoting more diverse representations during training. The contrastive learning loss function, based on InfoNCE, is formulated as follows.

\[
L_{{con}} = -\frac{1}{N} \sum_{i=1}^N \log \frac{\exp\left(\frac{{sim}(\mathbf{z}_i, \mathbf{z}_{i+})}{\tau}\right)}{\exp\left(\frac{{sim}(\mathbf{z}_i, \mathbf{z}_{i+})}{\tau}\right) + \sum_{j=1}^K \exp\left(\frac{{sim}(\mathbf{z}_i, \mathbf{z}_{i-})}{\tau}\right)}
\]

where \(z_i\) and \(z_i^+\) form a positive pair, \(z_i\) and \(z_j^-\)  form negative pairs, \({sim}(\cdot, \cdot)\) denotes the similarity function between two samples, and \({\tau}\) is the temperature parameter that controls the smoothness of the it{softmax} function. Given the large pool of negative samples, we employ random sampling to select $K$ negative samples for computation.

\subsection{Spatial-temporal symmetry-aware graph transformer for basketball tactics modeling}

In this section, we introduce the spatial-temporal modeling for basketball tactics based on the Graph Transformer architecture, where each tactical expert operates independently with non-shared parameters. This module consists of an initial data embedding and positional encoding layer, stacked $L$ spatial-temporal encoder layers, and a representation fusion layer. It forms the core encoder for the \model, which mines spatial-temporal dependency patterns from historical data and learns a mapping function from historical embeddings to the future. Uniquely, the \model model incorporates geometric deep learning into the Graph Transformer architecture, modeling the court through four second-order dihedral symmetry views. Specifically, we propose a group-equivariant attention mechanism that processes these views simultaneously and allows them to interact explicitly without breaking the symmetry. The symmetry constraints on the message passing layers ensure that the output is consistent with the input transformation. Besides that, considering the significant instantaneous changes in rate and direction of actions such as breakthroughs, jump shots, steals, quick turns, defensive transitions, fast break layups, counterattacks, and rebounds in a basketball game, as well as the impact of passing actions directly related to ball possession on other player nodes with a time delay, which could be associated with factors like pass speed and distance, player reaction time, defensive interference, and tactical patterns, we need to explicitly model this delay effect to enhance the representation of nodes in discrete spatial slices. Therefore, \model incorporates the short-term delay effects feature transformation module into the spatial self-attention head, integrating multiple delay patterns obtained from k-Shape clustering into each node’s key value based on similarity weighting, and considering the delay effects of matching nodes through cross-attention mechanism.

\subsubsection{Initial spatial-temporal data embedding and positional encoding layer}

The initial spatial-temporal data embedding and positional encoding layer converts the input of player nodes in the spatial-temporal graph into a high-dimensional representation. First, the raw input X with the static features \texttt{id}, \texttt{team}, \texttt{position}, and dynamic features \texttt{head} collected from the player nodes is transformed into \(\mathbf{X}_{{data}} \in \mathbb{R}^{T \times N \times D}\) through a fully connected layer, where D is the embedding dimension. We further design a spatial-temporal embedding mechanism to incorporate the necessary knowledge into the model, including the data augmentation under symmetric views to perceive the \(D_2\) transformation events of the court, and the spatial graph Laplacian positional encoding to encode the structure of the player relation graph.

Symmetry-aware data augmentation. \model uses geometric deep learning to explicitly model the symmetric views of the basketball court \(D_2\), avoiding the need for the model to implicitly learn these symmetries from historical data. It assumes that when the \texttt{head}, \texttt{headOrientation}, \texttt{hips}, \texttt{foot1}, \texttt{foot2} and \texttt{speed\_mps} features in player node dynamics exhibit \(D_2\) group transformations, the predictions for downstream tasks are consistent. To generate high-dimensional embeddings for the four transformations: identity transformation, 180-degree rotation around the x-axis, 180-degree rotation around the y-axis, and 180-degree rotation around the z-axis, we first apply view-consistent flips to each feature in the player node’s historical dynamic representations, including \texttt{head}, \texttt{headOrientation}, \texttt{hips}, \texttt{foot1}, \texttt{foot2} and \texttt{speed\_mps}. The transformation matrix \(R_{axis}(\theta)\) for the configuration \(D_2 \), where \(\theta\) represents the rotation angle around the axis and \((a,b,c)\) is the unit vector of the rotation axis:

\[
R_{{axis}}(\theta) = 
\begin{bmatrix}
a^2 (1 - \cos \theta) + \cos \theta & ab (1 - \cos \theta) - c \sin \theta & ac (1 - \cos \theta) + b \sin \theta \\
ab (1 - \cos \theta) + c \sin \theta & b^2 (1 - \cos \theta) + \cos \theta & bc (1 - \cos \theta) - a \sin \theta \\
ac (1 - \cos \theta) - b \sin \theta & bc (1 - \cos \theta) + a \sin \theta & c^2 (1 - \cos \theta) + \cos \theta
\end{bmatrix}
\]

Data augmentation based on symmetric views can alleviate overfitting issues in complex Graph Transformer architecture and generate initial embeddings for the group-equivariant attention mechanism in the subsequent spatial-temporal dependency encoding layers. Specifically, we obtain four symmetric views of the data: \(\mathbf{X_i}, \mathbf{X_{\leftrightarrow}}, \mathbf{X_{\uparrow}}, \mathbf{X_{\downarrow}} \in \mathbb{R}^{T \times N \times D}\), where T denotes the time steps, N represents the number of nodes, and D is the original embedding dimension. These embeddings are then transformed through a simple fully connected layer to \( \mathbf{X}_i, \mathbf{X_{\leftrightarrow}}, \mathbf{X_{\uparrow}}, \mathbf{X_{\downarrow}} \in \mathbb{R}^{T \times N \times D}\), where d is the reduced embedding dimension. This transformation allows different dimensions of the initial representations to interact non-linearly in a high-dimensional space. The resulting representations are subsequently denoted as \(X_{data}\).

bf{Laplace eigenvector positional encoding}. The multihead self-attention mechanism in the Transformer can effectively expand the receptive field for message passing on graphs, alleviating the over-smoothing problem commonly encountered in GNN-based models, which often face challenges in capturing long-range dependencies across multiple hops. In conventional machine translation tasks, the Transformer models the input text sequence as a fully connected graph, facilitating message passing between any two nodes regardless of their position in the sequence. However, unlike text sequences, graphs are non-Euclidean structures that embed rich topological information. It is crucial to consider this topological context alongside node features during the representation learning process. Specifically, in modeling basketball scenarios where the graph comprises only 10 player nodes, performing spectral decomposition in the Fourier domain on the graph Laplacian does not introduce substantial computational overhead. Consequently, \model\ integrates Laplace eigenvector positional encoding to preserve the topological structure of the graph in the initial data embedding layer. We perform a spectral decomposition on the normalized Laplacian matrix.

\[
\Delta = I - D^{-1/2} A D^{-1/2}, \quad \Delta = U \Lambda U^T
\]

where A represents the adjacency matrix, D is the degree matrix, I denotes the identity matrix,  A is the diagonal matrix (with the diagonal elements being the Laplacian eigenvalues) and U is the matrix of Laplacian eigenvectors. 

In \model, we compute the cosine similarity between nodes based on the initial feature embeddings for \texttt{team}, \texttt{position}, and \texttt{head} to generate an adjacency matrix that represents the probability of pass between players. Since the adjacency matrix is only used for calculating the Laplacian eigenvector positional encoding, and the subsequent self-attention-based message passing does not follow the graph edges, this approach not only avoids affecting the subsequent player node representation updates but also reduces the number of parameters required during training. This helps mitigate the risk of overfitting in a small-domain dataset when using the complex, hyperparameter-heavy model based on the Graph Transformer architecture. The smaller eigenvalues of the Laplacian matrix correspond to low-frequency eigenvectors, which capture slowly varying features across the graph and better represent the global structural information. In contrast, the larger eigenvalues correspond to high-frequency eigenvectors that capture rapidly varying features, reflecting the structural differences between neighboring nodes and local noise in the graph. Therefore, we use the eigenvectors corresponding to the smallest nonzero k eigenvalues as a new set of bases in the spectral graph space to capture global structural information. By generating graph Laplacian embeddings through the linear projection of node representations onto this new set of bases, we can effectively preserve the global structure of the graph.

\[
\mathbf{U}_{{spe}} = \begin{bmatrix} u_1, u_2, \dots, u_k \end{bmatrix}, \quad \mathbf{X}_{{lap}} = \mathbf{X}_{{data}} \, \mathbf{U}_{{spe}}
\]

bf{Transformer positional encoding}. We also adopt the Transformer sinusoidal positional encoding, treating all player nodes on the spatial-temporal graph as a sequence to preserve interaction order information. Additionally, we apply normalization to enhance training robustness.

\[
\mathbf{X}_{{pe}_{2i}} = \sin\left( \frac{pos}{10000^{2i/d}} \right), \quad
\mathbf{X}_{{pe}_{2i+1}} = \cos\left( \frac{pos}{10000^{2i/d}} \right), \quad
\mathbf{X}_{{pe}_{{norm}}} = \frac{\mathbf{X}_{{pe}} - mean(\mathbf{X}_{{pe}})}{std(\mathbf{X}_{{pe}}) \times 10}
\]

The output of the initial data embedding and positional encoding layer is formed by simply summing the initial feature embedding, the Laplacian eigenvector positional encoding, and the Transformer positional encoding, simply denoted as $X$:

\[
X_{emb} = X_{data} + X_{lap} + X_{pe}
\]

\subsubsection{Spatial-temporal dependency encoding layer}

This section explains how the initial player node representations from the four views produced by the initial data embedding and positional encoding layer are passed through stacked L spatial-temporal dependency encoding layers. During this process, we aggregate and update representations of temporal and spatial information to generate high-dimensional latent space embeddings of player nodes. We first segment the tensor \(X \in R^{T \times N \times d}\) into temporal slices \(X_{t::} \in R^{N \times d}\) and spatial slices \(X_{:n:} \in R^{T \times d}\). The temporal slices are used for spatial attention head, modeling the spatial dependencies between player nodes at each observation time. The spatial slices are used for the temporal attention head, modeling the temporal dependencies of each player node.
Spatial delay-effect feature transformation module. In the spatial attention head, we design a delay-effect feature transformation module to allow the representation of player nodes at each observation time to capture delay pattern information within a time window. First, for each time slice t, we obtain the query, key, and value matrices:

\[
\mathbf{Q}_t^S = \mathbf{X}_{t::} \cdot \mathbf{W}_Q^S, \quad
\mathbf{K}_t^S = \mathbf{X}_{t::} \cdot \mathbf{W}_K^S, \quad
\mathbf{V}_t^S = \mathbf{X}_{t::} \cdot \mathbf{W}_V^S
\]

where \(W_{Q}^{S}\), \(W_{K}^{S}\), \(W_{V}^{S} \in R^{d \times d^{'}}\) are learnable parameter matrices, and \(d^{'}\) is the dimension of the query, key, and value matrices. 
Next, we identify a set of representative short-term delay-effect patterns in basketball from the spatial slices of all nodes. Specifically, we slice the historical tensor \(X \in R^{T \times N \times d}\) of all nodes using a sliding time window S, where each window slice has dimensions \(R^{S \times N \times d}\). Since k-Shape clustering can retain the original time series dimensions and is invariant to scaling and shifting, we apply k-Shape clustering to all the slices and extract the cluster centroids \(P = p_{i} \mid i \in [1, \dots, N_{p}]\) as a set of short-term delay-effect patterns, where \(p_{i} \in R^{S \times N \times d}\) and \(N_{p}\) represents the total number of extracted patterns. We then compare the historical representation sequence of each node over a time window S with the extracted patterns. By performing a weighted sum, the information from similar patterns is integrated into the key value of that node’s time slice, allowing the subsequent computation of attention scores to take into account the short-term delay effects of the query node.

Specifically, we used a time window of the same length S to extract the historical spatial-temporal representation sequence of the target node \(X_{(t-s+1),n}\). We then apply linear interactions between \(X_{(t-s+1),n}\) and \(p_i\) using \(W^{u}\) and \(W^{m}\), mapping them into a higher dimensional space. Here, \(W^{u}\) and \(W^{m}\) are learnable parameter matrices.

\[
u_{t,n} = \mathbf{X}_{(t-s+1:), n} \cdot \mathbf{W}^u, \quad
m_i = p_i \cdot W^m
\]

By calculating the inner product between \(u_{t,n}\) and \(m_{i}\), the historical representation sequence of the node is compared with the delay-effect patterns, yielding the attention score \(w_{i}\):

\[
w_i = {softmax}\left( u_{t,n}^i \cdot m_i \right)
\]

Subsequently, we perform a weighted sum of the set of delay effect patterns based on the attention scores.

\[
r_{t,n} = \sum_{i=1}^{N_p} w_i \cdot \left( p_i \cdot W^c \right)
\]

Where \(W^{c}\) is a learnable parameter matrix and \(r_{t,n}\) contains historical spatial-temporal information of the player node n from the time slice $t-S+1$ to $t$.
Finally, we integrate the historical spatial-temporal information of the player node into the key value of the time slice at $t$, allowing the query matrix of the target node to account for short-term information when calculating the attention scores, rather than relying solely on the information from the current time slice.

\[
k_{t,n}^S = k_{t,n}^S + r_{t,n}
\]

Spatial group-equivariant attention mechanism. \model introduces geometric deep learning to explicitly model symmetrical regularities in basketball data. Geometric deep learning imposes constraints on the fitting function by utilizing geometric priors when fitting the mapping function in the intermediate layers of the neural network. This ensures that when the input undergoes a certain geometric transformation, the output of the neural network exhibits predictable behavior. Mathematically, there exists a set of input transformations \(\{g_1, g_2, \dots, g_n\}\) belonging to the symmetry group $G$, under which the true labels \(y(X,E)\) in the dataset remain consistent, i.e.:

\[
y(g_1(X, E)) = y(g_2(X, E)) = \dots = y(g_n(X, E))
\]

To design a mechanism in the neural network that enforces the aforementioned G transformation consistency, given a predictor \(f_{g}^{inv}\), we ensure that all transformed versions have identical outputs by averaging the prediction results across all group-transformed inputs.

\[
f_g^{{inv}}(X, E, g) = \frac{1}{|\mathbb{G}|} \sum_{g \in \mathbb{G}} f_g\left( g(X), g(E), g(g) \right)
\]

Although this enforced mechanism intuitively constrains the consistency of input-output, it only feeds each symmetric view independently into the neural network predictor and constrains the outputs. It does not allow different symmetric views to interact within the network. This approach results in the distribution of gradients during backpropagation based on the weights of the learned parameters of the network. In basketball scenarios, where the symmetry group \(G = D_2 = \{ {id}, \leftrightarrow, \uparrow, \leftrightarrow \uparrow \}\) consists of only four transformations, it is feasible to design finer-grained interactions among these four symmetric views within the network. This allows the parameters in the network layers to directly reflect symmetry without having to implicitly learn it. In \model, we propose a group-equivariant attention mechanism that allows input views to interact during the self-attention-based message-passing stage in the Graph Transformer:

\[
H_{{spatial\_}g} = {Message\_Passing}_{\mathbb{G}}(X_g) = \frac{1}{|\mathbb{G}|} \sum_{h \in \mathbb{G}} {Message\_Passing}_{\mathbb{G}}\left( X_h \, \Vert \, X_{g^{-1} h} \right)
\]

Where \(H_{spatial\_g}\) represents the final output of view g encoded by the spatial attention head, \(\Vert\) denotes the concatenation operation, \(g^{-1}\) is the inverse transformation of g. In \(D_2\), all transformations are self-inverse, meaning \(g^{-1}=g\); \(g^{-1}h\) is the composition of two transformations, \(X_h\) represents the output of the spatial-temporal data embedding and position encoding layer after the h transformation.
Information transfer in the Graph Transformer architecture is based on the self-attention mechanism. We calculate the attention scores between all player nodes in each spatial slice to capture spatial dependencies. The attention scores in the spatial domain for the t-th time slice, based on the self-attention mechanism, are:

\[
A_t^S = \frac{(\mathbf{Q}_t^S)(\mathbf{k}_t^S)^\top}{\sqrt{d'}}
\]

It can be observed that \(A_{t}^{S} \in R^{N \times N}\) has different values in different time slices, thus capturing the dynamic spatial dependencies in basketball.
In the Graph Transformer, message passing occurs globally, enabling the capture of spatial dependency patterns even between distant nodes. We perform a weighted sum of the values based on the attention scores to obtain the output of the spatial self-attention head.

\[
H_{{spatial\_g}} = {Spatial\_Self\_Attention}\left( \mathbf{Q}_t^S, \mathbf{k}_t^S, \mathbf{V}_t^S \right) = {softmax}\left( A_t^S \right) \mathbf{V}_t^S
\]

It is worth noting that not only do we generate four symmetric views through \(D_{2}\) transformations in the initial spatial-temporal data embedding layer to augment the sparse, continuous spatial-temporal data in basketball, but the \model\ group-equivariant attention mechanism also allows these symmetric views to be input into the model simultaneously. Moreover, these views explicitly interact during the self-attention-based message-passing process, with symmetry being enforced throughout.

Temporal self-attention module. The temporal self-attention module models the temporal dependencies of the player nodes based on the spatial slice \(X_{:n:} \in R^{T \times d}\) of the initial spatio-temporal tensor X. Similar to the spatial self-attention module, we first obtain the query, key, and value matrices for each node n in the spatial slices for the temporal attention head:

\[
\mathbf{Q}_n^T = \mathbf{X}_{:n} \cdot \mathbf{W}_Q^T, \quad
\mathbf{K}_n^T = \mathbf{X}_{:n} \cdot \mathbf{W}_K^T, \quad
\mathbf{V}_n^T = \mathbf{X}_{:n} \cdot \mathbf{W}_V^T
\]

Here,  \(W_{Q}^{T}\), \(W_{K}^{T}\), \(W_{V}^{T} \in R^{d} \times d^{'})\) are learnable parameter matrices, and \(d^{'}\) represents the dimension of the query, key, and value matrices.

The attention score for player node n in the spatial slices, based on the self-attention mechanism in the temporal space, is:

\[
A_n^T = \frac{(\mathbf{Q}_n^T)(\mathbf{K}_n^T)^\top}{\sqrt{d'}}
\]

It can be observed that \(A_{n}^{T} \in R^{N \times N}\) has different values across different spatial slices, thereby capturing the varying temporal dependency patterns of different player nodes in basketball. Message passing in the Graph Transformer architecture occurs at any time point for each node, allowing it to capture long-term dependencies on player nodes. We perform a weighted sum of the values based on the attention scores to obtain the output of the temporal self-attention head:

\[
H_{{temporal}} = {Temporal\_Self\_Attention}\left( \mathbf{Q}_n^T, \mathbf{K}_n^T, \mathbf{V}_n^T \right) = {softmax}\left( A_n^T \right) \mathbf{V}_n^T
\]

\subsubsection{Spatial-temporal representation fusion layer}

We concatenate the embeddings encoded by the spatial attention head \(H_{spatial}\) and the temporal attention head \(H_{temporal}\), which serves as the output of the spatial-temporal dependency modeling module based on the Graph Transformer architecture for basketball tactics.

\[
H_{{ST-Embedding}} = H_{{spatial}} \, \Vert \, H_{{temporal}}
\]

\subsection{Graph-grounded large language model}
The embeddings encoded by the basketball tactic spatial-temporal modeling module can be directly utilized for downstream tasks using traditional single-step recurrent prediction or multi-step one-shot prediction method. However, in basketball scenarios, player rotations and team matchups frequently result in node distribution shifts compared to pretrained datasets. This necessitates that predictive model possess inductive biases and out-of-distribution generalization capabilities, enabling accurate predictions on unseen datasets and facilitating long-term system maintenance. Inspired by the emergent capabilities exhibited by large language models (LLMs) after large-scale text pre-training and the successful application of LLMs in building multi-modal models within the field of computer vision domain, we propose that the text modality currently offers the most effective medium for achieving zero-shot prediction on graphs. Nevertheless, graph representation learning and large language model inferring have fundamentally different downstream task implementations logic. Thus, the key challenge lies in enabling LLMs to comprehend graph topology information to transfer their capabilities to graph tasks. Although LLMs can gain a basic understanding of graph structures through prompt engineering like chain-of-thought (CoT) reasoning, they still fail the two-hop Weisfeiler-Lehman test, indicating that their understanding of graph structures is far inferior to traditional GNN-based graph encoders. Drawing inspiration from LLaVA, which uses ViT as an image tokenizer, we continue to employ traditional graph encoder to encode graph structure information and map the embedding space to the text space of large models, enabling them to comprehend graph structure information.

\subsubsection{Two-stage semantic alignment}

\textbf{Text-aware spatial-temporal graph tokenizer for basketball tactics}. In the first stage, the spatial-temporal encoder aligns the encoded information with the text-encoded information of the spatial-temporal nodes using CLIP, ensuring that the embeddings, guided by the semantics of the text, capture relevant information. This allows the node information to be retained in the embedding vectors in semantic form, reducing the training difficulty for subsequent alignment with the text space of the large model using a linear layer, and minimizing information loss during alignment. It is worth noting that the CLIP model adopts a bidirectional training strategy, setting loss functions for both text-to-node embedding alignment and node-to-text embedding alignment, enabling bidirectional optimization. This ensures that the model can effectively retrieve content from both modalities, whether the input is node embeddings or text, thereby supporting a wider range of more complex downstream tasks.

\textbf{Alignment of graph and text token representation spaces}. In the second stage, the node embeddings from CLIP contrastive learning are aligned with the large model’s text space via a linear projector, enabling the model to process \texttt{<graph>} tokens as it would text tokens, thus acquiring an understanding of graph structures. In \model, we use a linear projector pretrained by GraphGPT on a large graph instruction-tuning dataset to convert the contrastively learned embeddings into the large model's word vector space. In GraphGPT, \texttt{<graph>} tokens are added to the model’s vocabulary, and through multitask supervised instruction fine-tuning, \texttt{<graph>} tokens are optimized to achieve maximum posterior probability representation in the large model's space when generating sentences from the dataset. During optimization, the gradients are backpropagated to the linear layer parameters to achieve alignment between the two modalities.

\subsubsection{Spatial-temporal in-context learning prompt strategy in basketball scenarios}

The generation architecture of large models' decoders, along with the principle of maximum a posteriori (MAP) generation, determines that reasoning requires providing contextual information. By designing fine-tuned instructions for different downstream tasks, we can maximize the activation of the knowledge learned during pre-training. Instruction design based on contextual learning consists of three parts: (1) prompt information; (2) problem description; (3) output format. Due to the relatively small parameter size of the base model used (7B), the quality of responses is sensitive to the prompt structure. Therefore, following the approach of GraphGPT, \model includes the instruction "Please analyze the provided time and player information before generating predictions. Think through the classification step by step to avoid incorrect associations." This guides the model to divide the final task into smaller, sequential tasks, enabling step-by-step reasoning to reach conclusions. Additionally, the large model performs poorly when generating absolute data, such as calculating the passing or shooting probabilities of a player in possession. However, it excels at ranking potential outcomes, such as identifying which player is most likely to have the ball or receive the ball, and performing binary judgment tasks like determining whether a player is in possession or is about to shoot. Therefore, our instruction design includes prompts such as "Rank from most likely to least likely" and "Answer with yes or no" to allow the model to focus on tasks where it excels, such as ranking and making judgments.

\begin{table*}[t]
\centering
\caption{Spatial-temporal in-context learning prompt design.}
\label{tab:prompt}
\small
\begin{tabularx}{\textwidth}{X X}

\toprule
\textbf{Node Classification Task:} Who is most likely to hold the ball at a specific moment? \\
\midrule

\textbf{Input:} Given a Graph Transformer encoding graph with \{20\} historical consecutive time slices: 
[\{timestamp\} $\langle$graph$\rangle$, \{timestamp\} $\langle$graph$\rangle$, ...], 
with player node information on the field described as 
[\{timestamp\} \{player\_id\} \{player\_information\} ...]. \\
\midrule

\textbf{Question:} Starting from the last time slice, who is most likely to have possession in the next \{time step\} time slices? 
Arrange in order from most likely to least likely, in the format ``\{player\_id\}'' separated by commas. 
Please answer strictly according to the answer template. 
\textcolor{orange}{Analyze the given time and player information thoroughly to generate the prediction. 
Follow a step-by-step approach to avoid incorrect associations.} \\ 
\midrule

\textbf{\model\ Response:} 
\texttt{Based on the provided information, \{player\_id\}, \{player\_id\}, \{player\_id\}, ... \{player\_id\}.} \\ 

\midrule
\textbf{Link Prediction Task}: Will the current ball handler pass the ball? To whom is s/he most likely to pass? \\
\midrule

\textbf{Input:} Given the Graph Transformer encoded graphs for \{20\} time slices: [\{timestamp\} $\langle$graph$\rangle$, ...], and player node information on the field described as [\{timestamp\} \{player\_id\} \{player\_information\} ...]. \\
\midrule

\textbf{Question:} Starting from the last time slice, will the player with the highest probability of holding the ball pass it? If so, to whom? Rank from most likely to least likely, in the format ``\{yes\} or \{no\}, \{player\_id\}'', separated by commas. \textcolor{orange}{Analyze the time slice based on the provided time and player information, then generate predictions. Think step by step to avoid incorrect assumptions.} \\
\midrule

\textbf{\model\ Response:} 
\texttt{Based on the provided information, \{yes\}, \{player\_id\}, \{player\_id\}, \{player\_id\}, ..., \{player\_id\}.} \\
\midrule

\textbf{Graph Classification Task}: Will the ball handler attempt a shot at a specific moment? \\
\midrule

\textbf{Input:} Given the Graph Transformer encoded graph at \{timestamp\}: $\langle$graph$\rangle$, and player node information on the field described as [\{player\_id\} \{player\_node\_information\}, ...]. \\
\midrule

\textbf{Question:} tarting from this time slice, will the player with the highest probability of holding the ball in the next \{time step\} time slices take a shot? Answer in the format ``\{yes\} or \{no\}``. \textcolor{orange}{Analyze the time slice based on the provided time and player information, then generate predictions. Think step by step to avoid incorrect assumptions.} \\ 
\midrule

\textbf{\model\ Response:} 
\texttt{Based on the provided information, \{yes\}.} \\ 

\bottomrule

\end{tabularx}
\vspace{-0.2in}
\end{table*}

\subsection{Two-stage pre-training strategy}

\textbf{Stage 1}. In the first stage, training is limited to the encoder component without integrating the large language model. It includes a mixture of tactic experts routing mechanism and a spatial-temporal graph transformer for basketball tactics modeling. The training objective is to minimize the loss function by predicting the ball possession transition for each player at the next time step, aiming to capture patterns from historical spatial-temporal data and yield robust player spatial-temporal representations. This pretraining step focuses on capturing the underlying dynamics of the game to ensure the model understands how players move and interact over time in a tactical environment.

\textbf{Stage 2}. The second stage employs an end-to-end training process where the focus shifts to optimizing the parameters of the CLIP model and the mixture of tactic experts routing mechanism, with the Graph Transformer parameters in the encoder kept frozen. The loss functions are designed to address three specific prediction tasks: (1) node classification task: who is most likely to have possession at a given moment? (2) link prediction task: will the player with possession pass the ball at a given moment? (3) graph classification task: will the player with possession shoot the ball at a given moment? The primary objective of this stage is to optimize the model for task-based completion, where the model is fine-tuned to understand and generate accurate predictions about game scenarios based on natural language inputs. By integrating these task-specific losses, the model is not only trained to predict individual actions but also optimized to deliver meaningful and context-aware predictions, enhancing its ability to handle complex, multi-modal tasks involving both visual and textual information.

\section{Experiments}\label{sec:experiments}

The extensive experiments to evaluate the performance of our method are from the following research questions (RQs):
\begin{itemize}
    \item \textbf{RQ1}: What are the experimental details, efficiency, and hyperparameter selection strategies of TacticExpert?
    \item \textbf{RQ2}: How does TacticExpert perform on supervised downstream tasks?
    \item \textbf{RQ3}: What is the contribution of each key component of TacticExpert to the overall performance?
    \item \textbf{RQ4}: How interpretable is TacticExpert?
\end{itemize}

\subsection{Experimental settings (RQ1)}

\textbf{Datasets}. The dataset used in our experiments is the ``DeepSport Basketball-Instants Dataset'' from Kaggle, which includes spatial-temporal data from the 2019-2020 season playoffs of the French professional basketball league. This dataset comprises 1,780 high-resolution sequential images, along with 729 corresponding camera calibration files. The dataset consists of high-resolution raw image files and ground truth key-point data. These images were captured during 38 games using the Keemotion system installed in 15 stadiums. Five fixed cameras in each venue captured two sequential images every 3 seconds, with a 40ms delay between frames. In TacticExpert, gating weights and tactical expert routing are computed using the initial player positioning at \texttt{relative\_time} = 0(tactic start) from the image modality. The initial spatial-temporal player node representations are generated from 13 feature dimensions, including \texttt{id}, \texttt{team}, \texttt{position}, \texttt{relative\_time}, \texttt{head}, \texttt{headOrientation}, \texttt{hips}, \texttt{foot1}, \texttt{foot2}, \texttt{foot1\_at\_the\_ground}, \texttt{foot2\_at\_the\_ground}, \texttt{speed\_mps} and \texttt{has\_ball}. Detailed descriptions of these features are provided in Table \ref{tab:2}.

\begin{table}[t]
\centering
\caption{Player feature details.}
\resizebox{1\textwidth}{!}{
\begin{tabular}{>{\raggedright\arraybackslash}p{3.5cm} >{\ttfamily}p{2cm} p{9cm}}
    \toprule
    \textbf{Feature Name} & \textbf{Data Type} & \textbf{Feature Description} \\
    \midrule
    id & str & Unique player identifier \\
    team & bool & Team affiliation of the player \\
    position & str & Position of the player on the field \\
    relative\_time & str & Time relative to the start of the game \\
    head & float array & Position of the player's head in (x, y, z) coordinates \\
    headOrientation & float & Orientation of the player's head \\
    hips & float array & Position and rotation of the player's hips in (x, y) coordinates \\
    foot1 & float array & Position and rotation of the player's first foot in (x, y) coordinates \\
    foot2 & float array & Position and rotation of the player's second foot in (x, y) coordinates \\
    foot1\_at\_the\_ground & bool & Whether the player's first foot is on the ground \\
    foot2\_at\_the\_ground & bool & Whether the player's second foot is on the ground \\
    speed\_mps & float & Speed of the player in meters per second \\
    has\_ball & bool & Whether the player has possession of the ball \\
    \bottomrule
    \label{tab:2}
\end{tabular}
}
\end{table}

\begin{figure}[t]
    \centering
    \includegraphics[width=\columnwidth]{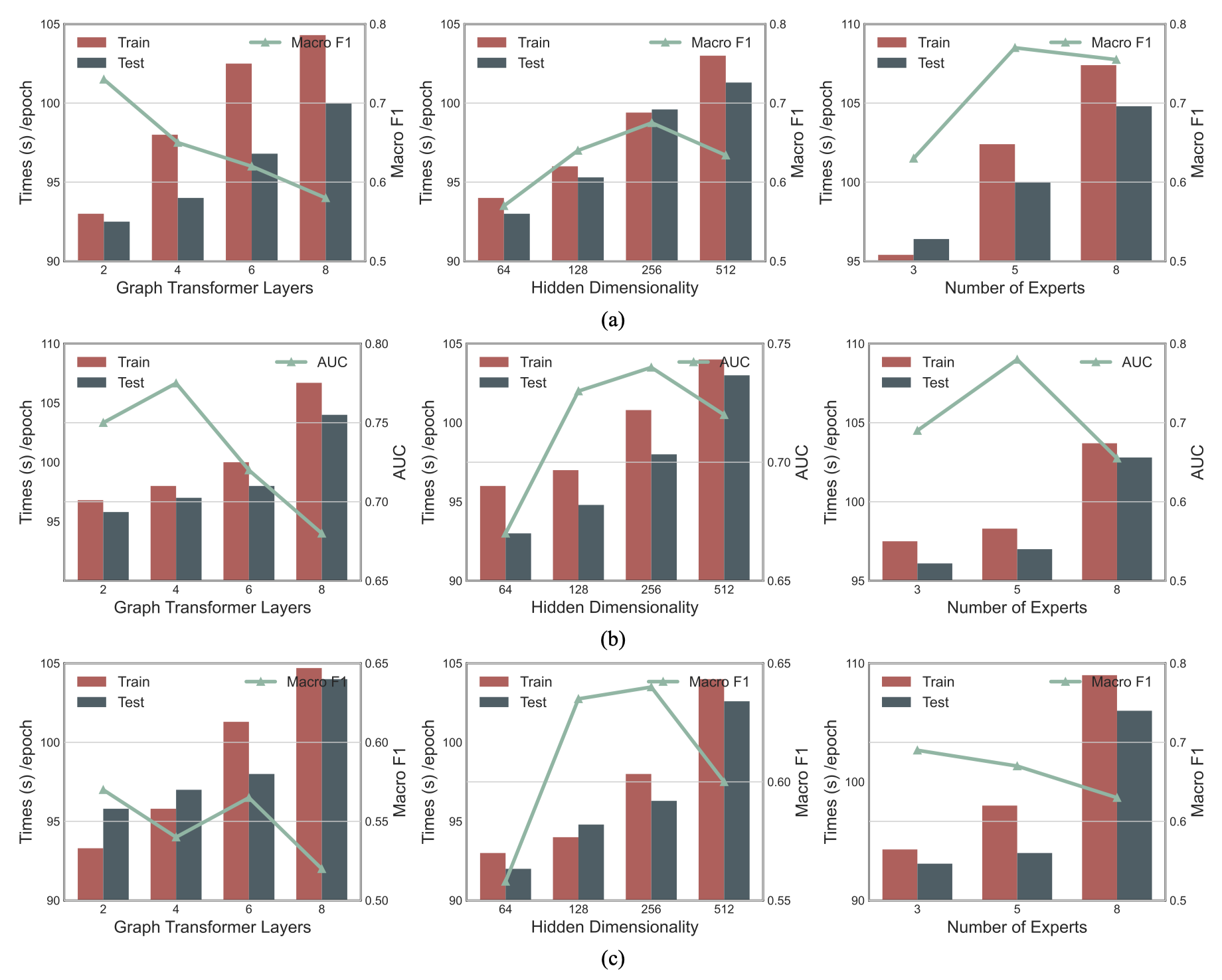}
    \caption{Performance, Efficiency, and Hyperparameter Selection of TacticExpert. (a) Node classification task; (b) Link prediction task; (c) Graph classification task}
    \label{fig:fig3}
\end{figure}

\textbf{Model settings and evaluation metrics}. We implemented TacticExpert primarily using PyTorch and Hugging Face Transformers libraries and employed FastChat as the framework to launch the base model Vicuna-7B-v1.5. The lightning version enabled training and inference on two NVIDIA RTX 3090 GPUs (24GB), with 50GB of memory. We used the AdamW optimizer for gradient backpropagation, with a batch size of 5 per GPU and a total of 200 epochs. Each training epoch took approximately 96 seconds, with inference times around 5 seconds. During Stage 1 training, the learning rate was set to 2e-5 for the initial 5 epochs to allow gradual expert adaptation and mitigate potential routing issues from random initialization. The learning rate was subsequently increased to 2e-3. In the experiments, we leveraged the pre-trained \texttt{llava-v1.6-mistral-7b-hf} model from the Hugging Face platform as a representation enhancer and the \texttt{all-MiniLM-L6-v2} model as the text encoder. We applied mean pooling of non-padding token embeddings in sentence-BERT. Hyperparameters for the graph encoder embedding dimension d were selected from {64, 128, 256, 512}, the number of spatial-temporal encoding layers L in the Graph Transformer from {2, 4, 6, 8}, and the number of experts from {3, 5, 8}. We used the Macro F1 score to evaluate node and graph classification tasks, and Area Under the Curve (AUC) to assess link prediction task. The best model was determined based on test set performance when validation scores peaked, with final predictions averaged over three runs. For the optimal model, we selected a 256-dimensional graph encoder embedding, a 2-layer spatial-temporal encoding architecture in the Graph Transformer, and 5 tactical experts (see Fig \ref{fig:fig3}). The choice of using two spatial-temporal encoding layers in the Graph Transformer warrants some discussion, as four layers yielded better results in the link prediction tasks. However, this decision was made based on two factors: (1) basketball spatial-temporal graphs are relatively small, consisting of only 10 player nodes with sparse connections, which risks overfitting with deeper models; (2) as noted in DIFFormer, stacking multiple attention layers in Graph Transformers provided limited benefit.

\subsection{Full-shot performance on supervised tasks (RQ2)}

We evaluated the performance of TacticExpert on three standard supervised downstream tasks: (i) node classification task “who is most likely to hold the ball at a specific moment?” (ii) link prediction task “will the current ball handler pass the ball? to whom is she/he most likely to pass?” (iii) graph classification task “will the ball handler attempt a shot at a specific moment?”. Table \ref{tab:3} summarizes the full-shot results for these tasks.

\begin{wraptable}{r}{0.6\textwidth}
\centering
\caption{Full-shot performance on supervised downstream tasks.}
\label{tab:3}
\resizebox{0.58\textwidth}{!}{%
\begin{tabular}{@{}l|cc@{}}
\toprule
\textbf{Supervised Downstream Tasks} & \textbf{Evaluation} & \textbf{Full-shot Performance} \\
\midrule
\textbf{Node Classification}    & Macro F1   & 0.8333  \\
\textbf{Link Prediction}   & AUC & 0.7264  \\
\textbf{Graph Classification}       & Macro F1  & 0.6750 \\
\bottomrule
\end{tabular}
}
\end{wraptable}

\subsection{Ablation study (RQ3)}

We first designed three ablation studies: w/o the mixture of tactical experts (-MoE), w/o the delayed-effect feature transformation module (-Delay), and w/o the group-equivariant attention mechanism (-Group), to assess the contributions of these core components to overall model performance. Additionally, we tested w/o Transformer positional encoding (-PE) and w/o Laplacian eigenvector encoding (-Lap) to evaluate their roles in capturing graph structure and interaction order. Lastly, we ran w/o CLIP alignment (-CLIP) to examine whether text-aware graph encoding can mitigate information loss while preserving semantics. The results of these ablation studies are shown in Table \ref{tab:4}.

\begin{figure}[h]
    \centering
    \includegraphics[width=\columnwidth]{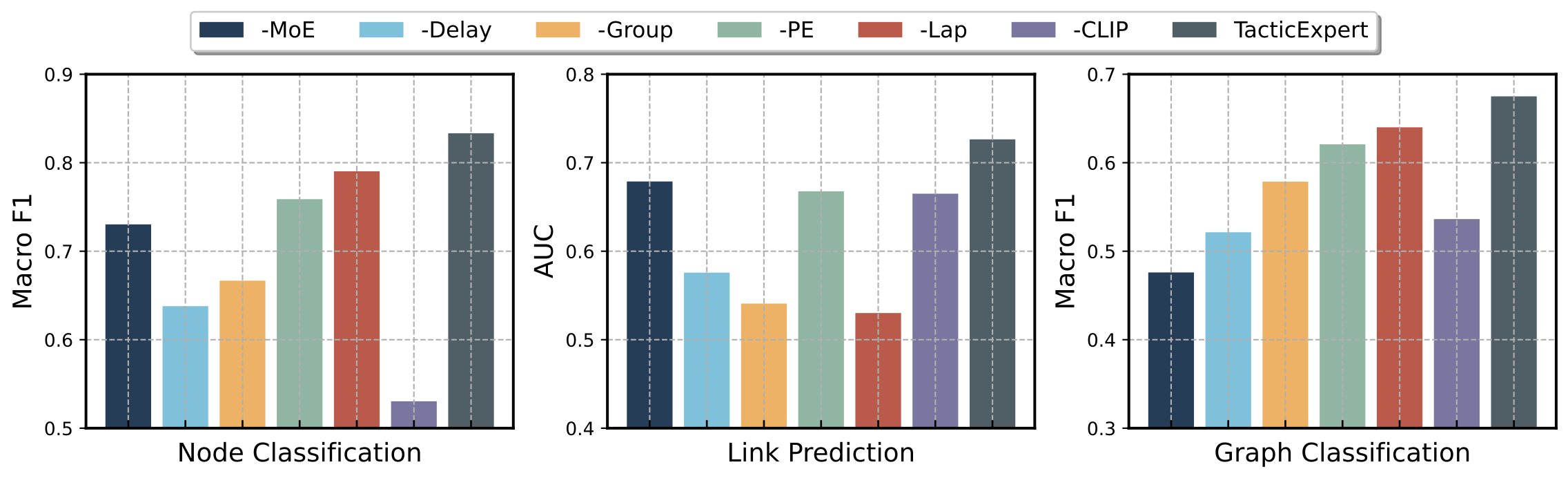}
    \caption{Ablation study on full-shot performance}
    \label{fig:fig4}
\end{figure}

\textbf{Ablation studies on core components}. In the ablation of the mixture of tactic experts module, we encoded all tactical information using a shared-parameter spatial-temporal Graph Transformer. In the ablation of the delayed-effect feature transformation module, the attention scores were computed using the key values derived from the transformed X. In the ablation of the group-equivariant attention mechanism, we deactivated symmetric embeddings across the four views in both the initial embedding and positional encoding layers, instead applying a weighted summation in the spatial attention head. As shown in Fig \ref{fig:fig4}, in the node classification task, the delayed-effect feature transformation and group-equivariant attention mechanism are pivotal to enhance the node representations by capturing spatial dependencies. In the link prediction task, these components remain crucial since effective link prediction relies on robust node representations and the relationships between connected nodes. In the graph classification task, the absence of the mixture of tactic experts module leads to model failure, suggesting that this module encodes high-level tactical knowledge essential for graph-wide analysis. Nonetheless, the delayed-effect feature transformation and group-equivariant attention continue to play key roles, as graph classification depends on global node information aggregated through the Readout function. 

\begin{table*}[t]
\scriptsize
\centering
\caption{Model variant performance on supervised downstream tasks}
\resizebox{1\textwidth}{!}{
\begin{tabular}{>{\raggedright\arraybackslash}p{3cm} >{\centering\arraybackslash}p{1.5cm} 
>{\centering\arraybackslash}p{1cm} >{\centering\arraybackslash}p{1cm} >{\centering\arraybackslash}p{1cm}
>{\centering\arraybackslash}p{1.1cm} >{\centering\arraybackslash}p{1cm} >{\centering\arraybackslash}p{1cm} >{\centering\arraybackslash}p{1cm}}
    \toprule
    \textbf{Supervised Downstream Tasks} & \textbf{Evaluation} & \textbf{Tactic-Expert} & \textbf{-MoE} & \textbf{-Delay} & \textbf{-Group} & \textbf{-PE} & \textbf{-Lap} & \textbf{-CLIP} \\
    \midrule
    Node Classification & Macro F1 & \textbf{0.8333} & 0.7303 & 0.6379 & 0.6667 & 0.7588 & \underline{0.7903} & 0.5304 \\
    Link Prediction & AUC & \textbf{0.7264} & \underline{0.6788} & 0.5758 & 0.5408 & 0.6677 & 0.5301 & 0.6650 \\
    Graph Classification & Macro F1 & \textbf{0.6750} & 0.4760 & 0.5214 & 0.5786 & 0.6208 & \underline{0.6400} & 0.5363 \\
    \bottomrule
    \label{tab:4}
\end{tabular}
}
\end{table*}

\textbf{Ablation Studies on positional encoding}. We conducted two ablation studies by disabling the Transformer’s sinusoidal positional encoding (-PE) and Laplacian eigenvector positional encoding (-Lap) to investigate how positional encoding affects the performance of TacticExpert by preserving graph structural information. As shown in Fig \ref{fig:fig4}, positional encoding has the most significant impact on the link prediction task, as this task requires the highest level of graph structural modeling. The absence of Laplacian eigenvector encoding causes the model to collapse in the link prediction task, indicating that this encoding captures node position information in the spectral view by preserving the graph’s topology as a signal. Additionally, we observed that sinusoidal positional encoding plays a more prominent role in node classification and graph classification tasks. This may be linked to the reliance on node representations in these tasks, as highlighted in the previous ablation study. In the key step of the attention mechanism, the sequence of interacting nodes requires positional encoding to influence attention scores, thereby impacting interaction strength.

\textbf{Ablation study on CLIP contrastive learning}. To examine the impact of aligning the graph embeddings with semantic information via CLIP contrastive learning, we conducted an ablation study by disabling the CLIP module. In this setup, graph embeddings are mapped directly through a linear layer to replace the \texttt{<graph>} token in the prompt. As shown in Fig \ref{fig:fig4}, the removal of CLIP contrastive learning leads to significant loss of node information in both node and graph classification tasks, which are more sensitive to node representations. CLIP enables the graph encoder text-aware and preserving graph structure in the semantic space. In contrast, the link prediction task, which emphasizes graph structure, is less affected by the absence of semantic information alignment through CLIP. Notably, the parameters of the linear projection layer in the ablation study were pre-trained with GraphGPT after CLIP contrastive learning, and the observed performance drop could also stem from the absence of fine-tuning these parameters.

\subsection{Interpretability analysis (RQ4)}
\begin{figure}[t]
    \centering
    \includegraphics[width=0.6\columnwidth]{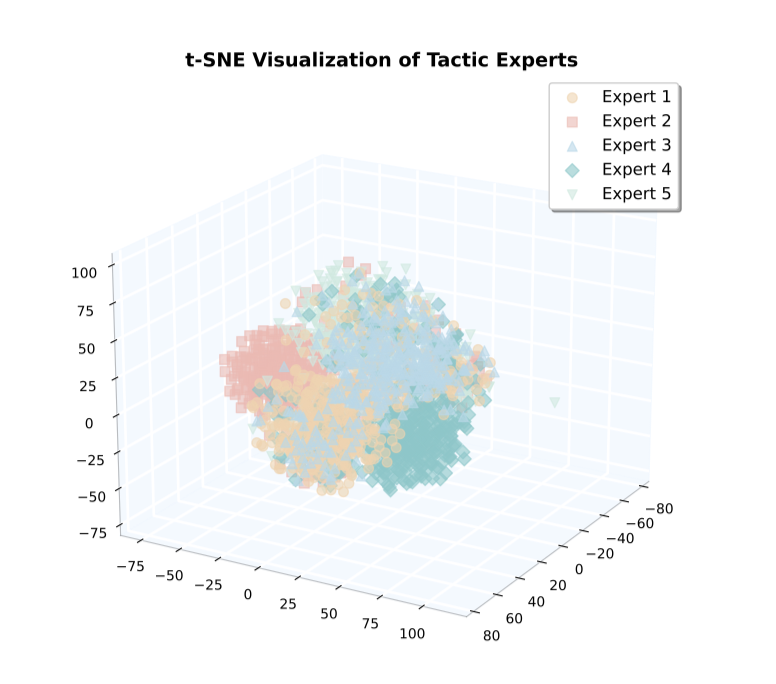}
    \caption{t-SNE visualization of tactic experts}
    \label{fig:fig5}
\end{figure}

\textbf{Do the five experts in the Mixture of Tactic Experts architecture specialize differently?} We treat the encoded embeddings from each expert as a sample pool. By randomly sampling from the five pools and applying t-distributed stochastic neighbor embedding (t-SNE) for dimensionality reduction, we visualize the embeddings in three-dimensional space. Fig \ref{fig:fig5} clearly shows that the embeddings of different experts naturally form distinct clusters, indicating that our semantic-based data partitioning for expert training, along with a contrastive learning mechanism that pulls embeddings of the same expert closer and pushes those of different experts apart, have enabled the experts to specialize in different tactical tasks. Notably, while the expert embeddings are distinguishable, they remain closely related in the 3D space. This suggests that we have found a balance between differentiating the experts’ roles and maintaining the core purpose of encoding player representations. The differences between experts are not so large as to hinder their shared task of player node encoding.

\begin{figure}[t]
    \centering
    \includegraphics[width=\columnwidth]{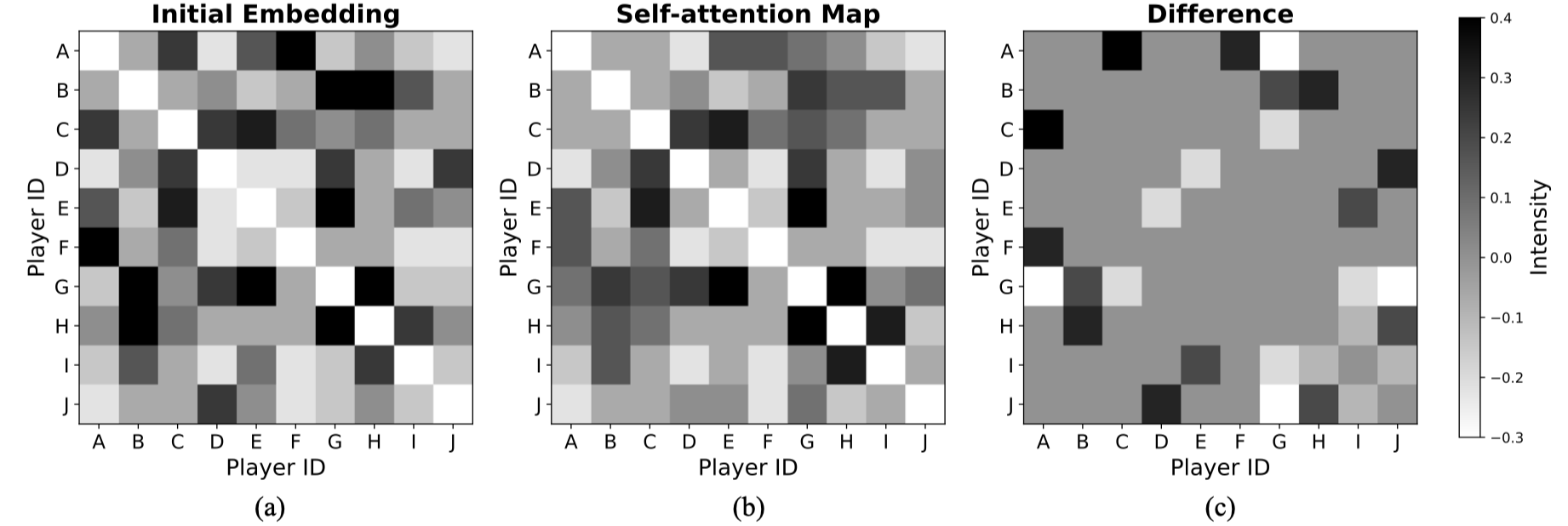}
    \caption{Visual analysis of long-range spatiotemporal dependencies. (a) Similarity of initial embeddings of player nodes; (b) Self-attention scores of player nodes; (c) Difference between a and b.}
    \label{fig:fig6}
\end{figure}

\textbf{Can TacticExpert capture long-range spatial-temporal dependencies?} We performed a visual analysis of spatial-temporal embeddings using a match segment from December 5, 2017, at LIMOGES stadium. Fig \ref{fig:fig6}.a illustrates the cosine similarity of player node embeddings at a specific time slice, indicating the initial correlations between nodes. Fig \ref{fig:fig6}.b presents the spatial attention scores from the spatial-temporal Graph Transformer module, showing the node correlations after spatial encoding. Fig \ref{fig:fig6}.c highlights the difference between (a) and (b). The black blocks in (c) represent node pairs with low initial correlation but high attention scores, indicating long-range spatial dependencies. TacticExpert computes attention scores across all nodes, transcending the limitations of traditional graph neural networks (GNNs), which confine information propagation to predefined edges. By doing so, the model effectively captures long-range spatial-temporal dependencies, allowing for more comprehensive pattern recognition and inference.

\newpage

\section{Conclusion}\label{sec:conclusion}

TacticExpert is the first approach to integrate spatial-temporal modeling of basketball tactics with large language models (LLMs), achieving training on unified pretraining tasks and inference across few-shot or even zero-shot scenarios and multiple downstream tasks. This breaks the constraint of the traditional end-to-end or pretraining + fine-tuning paradigm, where the training tasks and downstream tasks need to align, and enhances the model's generalization capability when there is a shift in node set type distribution or when downstream tasks differ from pretraining tasks. It also demonstrates strong inductive bias capabilities.

We designed a series of mechanisms for fine-grained modeling of spatial-temporal dependencies in basketball game scenarios, leading to high-quality player node representations enriched with spatial-temporal information:
\begin{itemize}
    \item The \textbf{Delayed Effect Feature Transformation Module} explicitly models the delayed effects during message passing between player nodes by preemptively identifying significant instantaneous changes in the rate and direction of target player nodes, as well as actions directly related to ball possession, such as passing and shooting. This enhances the representation of player nodes across time slices.  
    \item The \textbf{Group-equivariant Attention Mechanism} applies symmetric data augmentation and explicit symmetric interactions based on dihedral group transformations in basketball. By constraining network layers with symmetric priors, this enables the model to perceive symmetry.  
    \item The \textbf{Hybrid Tactical Expert Routing Mechanism} differentiates the modeling of various tactics, leveraging data from similar tactical patterns to enhance learning of spatial-temporal dependencies. This mitigates overfitting caused by uneven tactical distributions in experimental datasets and improves robustness.
\end{itemize}

The multi-head self-attention mechanism based on the Graph Transformer architecture allows TacticExpert to capture dynamic spatial-temporal relationships over long time horizons and long distances, overcoming the limitation of GNN models where information flow is restricted to edges. Furthermore, visualization analysis demonstrates the strong interpretability of the proposed model. The implementation code and datasets are open-sourced to promote community development.

Our work is not without its limitations. Due to the limited amount of training data, we were unable to retrain the linear layers aligned with the LLM space without overfitting. Instead, we used parameters from GraphGPT trained on the OGB-arXiv and PubMed datasets for our linear layers. If more training data becomes available, these parameters can be used for initialization to improve convergence speed.  Since LLMs rely on contextual learning, target player node information must be re-described in text form within the prompt. This may lead to performance degradation when working with graphs that lack textual information for nodes. Furthermore, assigning the entire task of graph structure encoding to the graph encoder and then aligning the encoded embedding space with the LLM space introduces a limitation in the utilization of graph structural information, as the encoding capacity is restricted to the graph encoder.

In the future, if larger basketball spatial-temporal datasets become available, we can explore more fine-grained LLM alignment methods, such as using the MoE (Mixture of Experts) architecture for the mapper. More powerful language models, such as Llama 3, could serve as the base model, providing higher-quality text token embeddings and stronger contextual learning capabilities. Instead of treating the LLM solely as a representation enhancer and downstream task predictor, additional Transformer layers could be added on top of the LLM embedding layer, enabling the entire LLM to act as a spatial-temporal encoder with the new Transformer layers being trainable. 

Additionally, experiments revealed that LLMs are not particularly sensitive to the quality of graph structural information embeddings. This raises the question of whether, in the era of large language models, we should persist in encoding multi-hop graph structural information or instead focus on tasks that can be accomplished using only one-hop neighbor information combined with prompt-based hints. 

\clearpage

\bibliographystyle{unsrtnat}
\bibliography{neurips_2024}

\end{document}